\documentclass[11pt]{article}
\usepackage{coling2020}
\usepackage{times}
\usepackage{url}
\usepackage{latexsym}

\usepackage{amssymb}
\usepackage{pifont}
\newcommand{\cmark}{\ding{52}}%
\newcommand{\xmark}{\ding{56}}%
\usepackage{subfigure}
\usepackage{bbding}
\usepackage{bm}
\usepackage{xcolor}
\usepackage{tabularx}
\usepackage{multicol}
\usepackage{multirow}
\usepackage{amsfonts}
\usepackage{soul}
\usepackage[utf8]{inputenc}
\usepackage{graphicx}
\usepackage{amsmath}
\usepackage{booktabs}
\usepackage{algorithm}
\usepackage{algorithmic}

\colingfinalcopy 

\title{Attention Transfer Network for Aspect-level Sentiment Classification}

\author{Fei Zhao$^*$ \hspace{1em} Zhen Wu$^*$ \hspace{1em} Xinyu Dai$^{\dagger}$\\
National Key Laboratory for Novel Software Technology, Nanjing University, China\\
Collaborative Innovation Center of Novel Software Technology and Industrialization, China\\
  {\tt \{zhaof, wuz\}@smail.nju.edu.cn, daixinyu@nju.edu.cn}
  \\}

\date{}

\begin{document}
\maketitle
\begin{abstract}
  Aspect-level sentiment classification (ASC) aims to detect the sentiment polarity of a given opinion target in a sentence. In neural network-based methods for ASC, most works employ the attention mechanism to capture the corresponding sentiment words of the opinion target, then aggregate them as evidence to infer the sentiment of the target. However, aspect-level datasets are all relatively small-scale due to the complexity of annotation. Data scarcity causes the attention mechanism sometimes to fail to focus on the corresponding sentiment words of the target, which finally weakens the performance of neural models. To address the issue, we propose a novel Attention Transfer Network (ATN) in this paper, which can successfully exploit attention knowledge from resource-rich document-level sentiment classification datasets to improve the attention capability of the aspect-level sentiment classification task. In the ATN model, we design two different methods to transfer attention knowledge and conduct experiments on two ASC benchmark datasets. Extensive experimental results show that our methods consistently outperform state-of-the-art works. Further analysis also validates the effectiveness of ATN. Our code and dataset are available at \url{https://github.com/1429904852/ATN}.
\end{abstract}

\section{Introduction}
\label{intro}

%
%
\blfootnote{
    %
    %
    \hspace{-0.65cm}  
    $^*$ Authors contributed equally.\\
    $^{\dagger}$ Corresponding author.\\
    This work is licensed under a Creative Commons Attribution 4.0 International Licence. Licence details: \url{http://creativecommons.org/licenses/by/4.0/}.
    %
    %
    %
    %
}

Aspect-level sentiment classification (ASC) is a fundamental task in sentiment analysis ~\cite{pang2008opinion,liu2012sentiment,Pontiki2014SemEval2014T4}, which aims to infer the sentiment polarity (e.g. positive, neutral, negative) of a given opinion target in a review sentence. An opinion target, also known as aspect term, refers to a word or a phrase in review describing an aspect of an entity. For example, the sentence ``\emph{The \textbf{tastes} are great, but the \textbf{service} is dreadful}'' consists of two opinion targets, namely ``\emph{tastes}'' and ``\emph{service}''. User's sentiment towards the opinion target ``\emph{tastes}'' is positive while negative in terms of target ``\emph{service}''. Traditional methods usually focus on designing a set of features such as bag-of-words or sentiment lexicon to train a classifier (e.g., SVM) for ASC~\cite{jiang2011target,kiritchenko2014nrc}. Motivated by the great success of deep learning in computer vision~\cite{krizhevsky2012imagenet}, speech recognition~\cite{DBLP:journals/taslp/DahlYDA12} and natural language processing~\cite{bengio2003neural}, recent works use neural networks to  learn low-dimensional and continuous text representations without any feature engineering, and achieve competitive results on the ASC task~\cite{Tang2016EffectiveLF}.

From the above example, we can see that a sentence sometimes refers to several opinion targets and they may express different sentiment polarities, thus one main challenge of ASC is to separate different opinion contexts for different targets. To this end, abundant state-of-the-art works employ attention mechanism~\cite{bahdanau2014neural} to capture sentiment words related to the given target, and then aggregate them to make sentiment prediction~\cite{wang2016attention,Tang2016AspectLS,ma2017interactive,chen2017recurrent,majumder2018iarm,fan2018multi}. Despite the effectiveness of attention mechanism, we argue that it fails to reach the full potential due to the limited ASC labeled data. It is well-known that the promising results of deep learning heavily rely on sufficient training data. However, the annotation of ASC data is very labour-intensive and expensive in real-world scenarios, because annotators need to not only identify all opinion targets in a sentence but also determine their corresponding sentiment polarity. The difficulty of annotation leads to that existing public aspect-level datasets are all relatively small-scale, which finally limits the potential of attention mechanism.

Despite the lack of ASC data, enormous labeled data of document-level sentiment classification (DSC) are available at online review sites such as Amazon and Yelp. These reviews contain substantial sentiment knowledge and semantic patterns. Therefore, one meaningful but challenging research question is how to leverage resource-rich DSC data to improve the low-resource task ASC. For this purpose,~\newcite{he2018exploiting} design the PRET+MULT framework to transfer sentiment knowledge from DSC data to ASC task through sharing shallow embedding and LSTM layer. Inspired by the capsule network~\cite{sabour2017dynamic},~\newcite{chen2019transfer} propose TransCap to share bottom three capsule layers, then separate two tasks only in the last ClassCap layer. Fundamentally, PRET+MULT and Transcap improve ASC by sharing parameters and multi-task learning, but they cannot accurately control and interpret what knowledge to be transferred. In this work, we directly focus on the aforementioned attention issue in the ASC task and propose a novel framework, \textbf{A}ttention \textbf{T}ransfer \textbf{N}etwork (ATN), to explicitly transfer attention knowledge from the DSC task for improving the attention capability of the ASC task. Compared with PRET+MULT and Transcap, our model achieves better results and retains good interpretability.

In the ATN framework, we adopt two attention-based BiLSTM networks, respectively, as the DSC module and base ASC module, and propose two different methods to transfer attention from DSC to ASC. The first transfer approach is called \emph{Attention Guidance}. Specifically, we first pre-train an attention-based BiLSTM on large-scale DSC data, then exploit the attention weights from the DSC module as a learning signal to guide the ASC module to capture sentiment clues more accurately, thereby acheiving improvements. The second approach adopts the way of \emph{Attention Fusion}, and directly incorporates the attention weights of the DSC module into the ASC module. The two approaches work in different ways and have their different advantages. \emph{Attention Guidance} aims to learn the attention ability of the DSC module and has faster inference speed, since it does not use external attention from DSC during the testing stage. In contrast, \emph{Attention Fusion} can leverage the attention knowledge of the DSC module during the testing stage and make more comprehensive predictions.

We conduct experiments on two benchmark datasets to evaluate different methods. The results indicate that the ATN model can be substantially improved by incorporating the two attention transfer approaches, and outperforms all compared methods on the ASC task.


\section{Model}
\begin{figure}[!htbp]
	\centering 
	\includegraphics[width=0.88\textwidth]{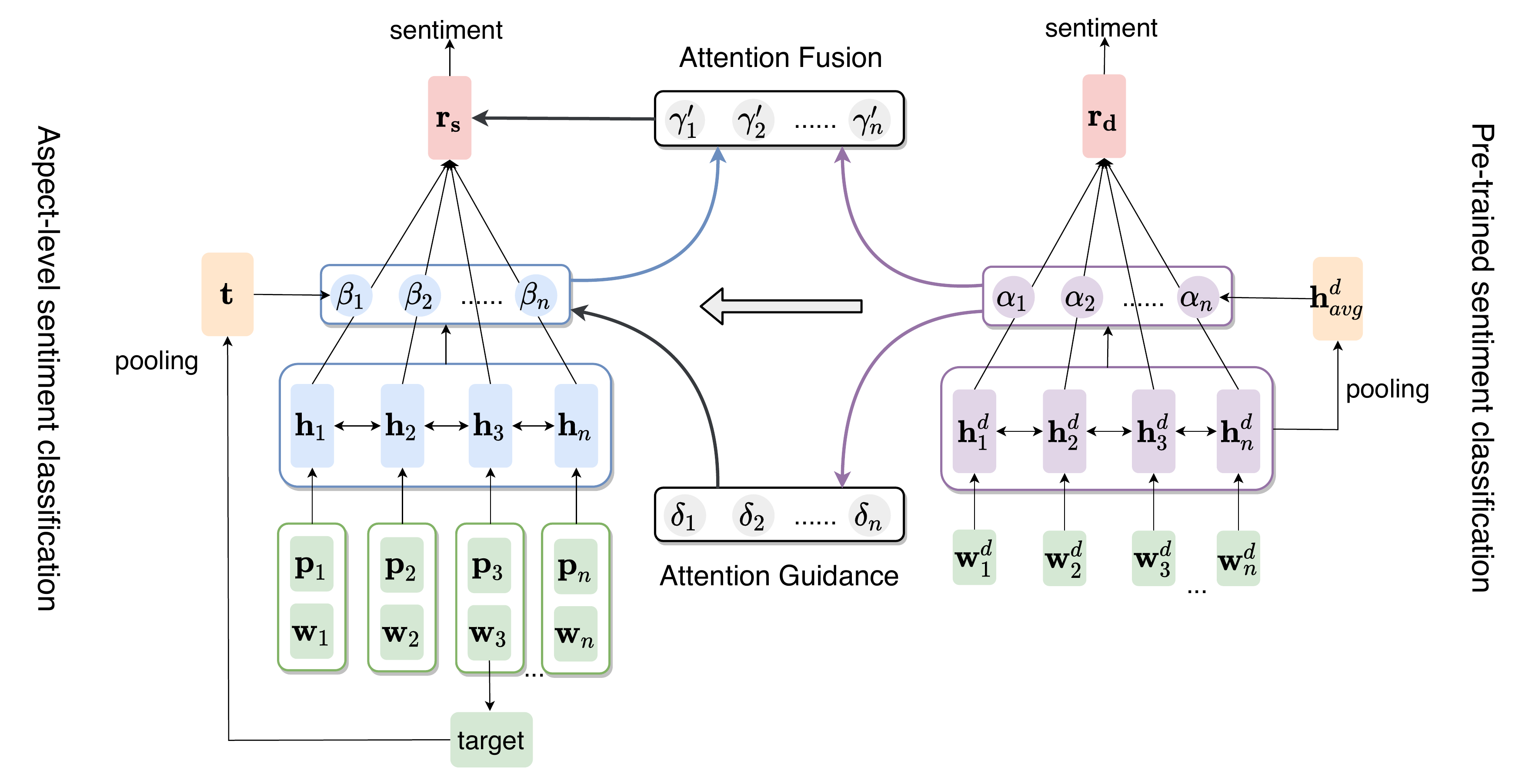}
	\caption{An illustration of our attention transfer network. The left one is the aspect-level sentiment classification, the right one is the pre-trained DSC module, and the middle part presents two proposed attention transfer approaches.}
	\label{fig:figure1label}
\end{figure}

Figure~\ref{fig:figure1label} shows the overall architecture of the Attention Transfer Network (ATN). It mainly consists of four parts:  the pre-trained DSC module, the base ASC module, and two attention transfer approaches. In this section, we will first give the task formalization of ASC and DSC, then introduce the attention-based pre-trained DSC module and base ASC module. Finally, we present the details of our proposed two attention transfer approaches, namely \emph{Attention Guidance} and \emph{Attention Fusion}.

\subsection{Task Formalization}
\textbf{ASC Formalization} Formally, given a sample $<s, t>$ from the ASC dataset $\mathcal{A}$, $s=\{w_1, w_2, ..., w_n\}$ is a review sentence consisting of $n$ words and $t=\{w_l, w_{l+1}, ..., w_r\}$ is a given opinion target containing $|r-l|$ words. The opinion target $t$ is a continuous subsequence of $s$. The goal of ASC is to predict the sentiment polarity (i.e., positive, neutral and negative) of the opinion target $t$ in the sentence $s$.

\noindent \textbf{DSC Formalization} For a review document $d$ from the DSC dataset $\mathcal{D}$, we regard it as a special long sentence $\{w_1^d, w_2^d, ..., w_n^d\}$ consisting of $n$ words. DSC aims to determine the overall sentiment polarity of the review document $d$.

\subsection{Pre-trainig DSC Module}
Before transferring attention knowledge, we first pre-train a DSC module on the large-scale DSC dataset $\mathcal{D}$. In this work, we employ a conventional attention-based BiLSTM as our DSC module.

For a review document $d=\{w_1^d, w_2^d, ..., w_n^d\}$, we map it into the corresponding word representations $\{\mathbf{w}_1^d, \mathbf{w}_2^d, ..., \mathbf{w}_n^d\}$ by looking up an embedding table $\mathbf{E}_{emb} \in \mathbb{R}^{|V|\times d_e} $, where $|V|$ is the vocabulary size and $d_e$ denotes the word embedding dimension. Then a BiLSTM network is applied to capture the contextual information for each word and generate a sequence of hidden states $\{\mathbf{h}_1^d, \mathbf{h}_2^d, ..., \mathbf{h}_n^d\}$. To  obtain the document representation $\mathbf{r}_d$, we employ the attention mechanism to aggregate the sentiment words that are significant for sentiment classification as follows:
\begin{equation}
\mathbf{r}_d = {\sum_{i=1}^n} \alpha_i {\bf h}_{i}^d,
\end{equation}
where $\alpha_i$ is the attention weight of $\mathbf{h}_i^d$ and defined as:
\begin{align}
\alpha_i &= \frac{\text{exp}(f({\bf h}_i^d, {\bf h}_{avg}^d))}{\sum_{j=1}^n \text{exp}(f({\bf h}_j^d, {\bf h}_{avg}^d))}, \label{dscatt} \\
f({\bf h}_i^d, {\bf h}_{avg}^d) &= {\bf h}_i^d \cdot {\bf W}_{d} \cdot {\bf h}_{avg}^d + {\bf b}_{d},
\end{align}
where $\mathbf{h}_{avg}^d$ is the average of all the hidden states, i.e., $\mathbf{h}_{avg}^d=\sum_{i=1}^{n}\mathbf{h}_i^d/n$, ${\bf W}_{d}$ and ${\bf b}_{d}$ are respectively the weight matrix and bias.

Finally, the representation $\mathbf{r}_d$ is fed to a linear layer and a softmax layer to predict the sentiment label of the review document $d$. We pre-train the DSC module by minimizing the cross-entropy loss between the predicted sentiment distribution and the ground truth. After pre-training is finished, all parameters in the DSC module are fixed.

\subsection{Base ASC Module}
As shown in the left part of Figure~\ref{fig:figure1label}, the base ASC module has a similar architecture to the DSC module. The difference is that the ASC task needs to model opinion target information. To obtain target-aware context representations, we additionally employ position embedding besides word embedding, which is an effective method of modeling position information~\cite{lin2016neural,gehring2016convolutional}. Therefore, the base ASC module is an attention-based BiLSTM network enhanced with position embedding.

Specifically, given a sentence $s=\{w_1, w_2, ..., w_n\}$ and an opinion target $t=\{w_l, w_{l+1}, ..., w_r\}$ in $s$, we first map each word $w_i$ into its word embedding representation $\mathbf{w}_i$ by using the word embedding table. To incorporate opinion target information with position embedding, we calculate the relative distance $l_i$ of each word $w_i$ to the opinion target $t$:
\begin{equation}
l_i=\begin{cases}
l-i & \text{ if } i <  l, \\ 
0 & \text{ if } l \leq   i \leq  r, \\ 
i-r & \text{ otherwise }. 
\end{cases}
\label{reldis}
\end{equation}
The distance index $l_i$ is mapped into the positional representation $\mathbf{p}_i$ by looking up a position embedding table $\mathbf{E}_{pos} \in \mathbb{R}^{L\times d_p}$, where $L$ denotes the maximal position index and $d_p$ is the embedding dimension. Then we concatenate the word embedding representation $\mathbf{w}_i$ and position embedding representation $\mathbf{p}_i$ as the repsentation $\mathbf{e}_i$ of the word $w_i$, i.e., $\mathbf{e}_i=[\mathbf{w}_i; \mathbf{p}_i]$, where $[\cdot; \cdot]$ denotes the vector concatenation operation. Similarly, we employ a BiLSTM to receive the word represenations $\{\mathbf{e}_1, \mathbf{e}_2, \cdots, \mathbf{e}_n\}$ as input and generate target-aware context representations $\{\mathbf{h}_1, \mathbf{h}_2, \cdots, \mathbf{h}_n\}$. Different from the attention part of the DSC module, we use the opinion target represenation $\mathbf{t}=\sum_{i=l}^{r} \mathbf{h}_i / (r-l)$ as query in the ASC task to extract target-dependent sentiment clues:
\begin{align}
f({\bf h}_i, {\bf t}) &= {\bf h}_i \cdot {\bf W}_{s} \cdot {\bf t} + {\bf b}_{s}, \\
\beta_i &= \frac{\text{exp}(f({\bf h}_i, {\bf t}))}{\sum_{j=1}^n \text{exp}(f({\bf h}_j, {\bf t}))},\label{ascatt}\\
\mathbf{r}_s &= {\sum_{i=1}^n} \beta_i {\bf h}_{i} \label{ascagg},
\end{align}
where ${\bf W}_{s}$ and ${\bf b}_{s}$ are respectively the weight matrix and bias.

Finally, the target-dependent sentence representation $\mathbf{r}_s$ is used for detecting the sentiment polarity of the target $t$, and the base ASC module can optimized by minimizing the following cross-entropy loss:
\begin{align}
\hat{y}_i = {\rm softmax}({\bf W}_{o} {\bf r}_s + {\bf b}_o),\\
\mathcal{L}_o = - \sum_{i \in \mathcal{A}}y_i log(\hat{y}_i)\label{ce},
\end{align}
where ${\hat y}_i$ and $y_i$ respectively are the predictive class distribution and golden class distribution.

\subsection{Attention Guidance}
To leverage the attention knowledge of the DSC module, we simultaneously input the sentence $s$ into the base ASC module and the pre-trained DSC module when performing the ASC task, generating the attention weights $\beta_i$ in Equation~\ref{ascatt} and $\alpha_i$ in Equation~\ref{dscatt}.

As mentioned before, the attention mechanism of the ASC module cannot reach full potential due to limited training data, which means that the attention weights $\beta_i$ may fail to capture target-relevant sentiment words. In contrast, sufficient DSC data enables the DSC module to extract sentiment words more accurately. Thus we propose the \emph{Attention Guidance} approach to guide the learning of the attention weights $\beta_i$ with the help of $\alpha_i$. Nevertheless, there is a tiny gap between the attention weights $\alpha_i$ and $\beta_i$. Since the DSC task only detects the overall sentiment of a review, the sentiment words captured by $\alpha_i$ are global and target-irrelevant. To make up the gap, we use a heuristic method to transform target-irrelevant attention weight $\alpha_i$ into target-relevant weight $\delta_i$:
\begin{align}
\alpha_i^\prime &= \frac{1}{2^{(l_i-1)}} \alpha_i, \\
\delta_i &= \frac{e^{\alpha_i^\prime}}{\sum_{i=1}^n e^{\alpha_i^\prime},}
\end{align}
where $l_i$ denotes the relative distance between the word and the target as in Equation~\ref{reldis}. We can see that a word nearer to the target receives a higher attention weight according to $\delta_i$, because the closer word has a bigger probability of modifier relation to the target.

Finally, we apply KL (Kullback–Leibler divergence) to describe the differences between attention distributions $\beta$ and $\delta$:
\begin{align}
KL(\delta || \beta) &= \sum_{i=1}^{n} \delta_i log\frac{\delta_i}{\beta_i}, \\
&= \sum_{i=1}^{n} (\delta_i log\delta_i-\delta_i log\beta_i) \label{klloss}.
\end{align}
In the pre-trained DSC module, the above term $\sum_{i=1}^{n}\delta_i log\delta_i$ in Equation~\ref{klloss} is invariant for the given sentence $s$ and the opinion target $t$. Therefore, we can minimize the loss $\mathcal{L}_a=\sum_{i=1}^{n}-\delta_i log\beta_i$ to guide the ASC module to focus on target-relevant sentiment words. In the \emph{Attention Guidance} approach, the final loss is defined as follows:
\begin{equation}
\mathcal{L} = \mathcal{L}_o + \lambda \mathcal{L}_a 
\label{lambda}.
\end{equation}
where $\lambda$ is the hyperparameter that controls the importance of $\mathcal{L}_a$ .

\subsection{Attention Fusion}
\emph{Attention Guidance} learns the attention ability of the DSC module through an auxiliary supervision signal. However, it cannot leverage the attention weights from the DSC module during the testing stage and wastes the pre-trained knowledge. To make full use of the additional attention capacity, we further propose the \emph{Attention Fusion} approach to incorporate them directly.

Specifically, we design a fusion gate $g$ to integrate the global attention weight $\alpha_i$ from the DSC module and the target-dependent attention weight $\beta_i$ from the ASC module, thereby generating more comprehensive and accurate attention weight $\gamma_i^\prime$:
\begin{align}
g &= \sigma({\bf W}_g[\alpha_i; \beta_i]), \\
\gamma_i &= g \alpha_i + (1-g)\beta_i, \\
\gamma_i^\prime &= \frac{e^{\gamma_i}}{\sum_{i=1}^n e^{\gamma_i}},
\end{align}
where $\sigma$ denotes sigmoid function and ${\bf W}_g$ is the weight matrix.

Finally, we replace $\beta_i$ in Equation~\ref{ascagg} with the new attention weight $\gamma_i^\prime$ to obtain the target-dependent sentence representation $\mathbf{r}_s$ for sentiment prediction.

\section{Experiments}
\subsection{Datasets and Metrics}
We evaluate our model on two ASC benchmark datasets from SemEval 2014 Task 4 \cite{Pontiki2014SemEval2014T4}. They respectively contain reviews from \emph{Restaurant} and \emph{Laptop} domains. Following previous studies~\cite{Tang2016AspectLS,chen2017recurrent,he2018exploiting}, we remove samples with conflicting polarities in all datasets. The statistics of the ASC datasets are shown in Table~\ref{table:table1}.

To pre-train the DSC module, we employ two larget-scale DSC datasets, respectively \emph{Yelp Review} and \emph{Amazon Review}~\cite{li2018delete}. The DSC dataset \emph{Yelp Review} is applied to transfer attention knowledge for the ASC dataset \emph{Restaurant}. The \emph{Amazon Review} is used for the dataset \emph{Laptop}. Table~\ref{table:table2} shows their statistics. In this work, we adopt Accuracy and Macro-F1 score as the metrics to evaluate the performance of different methods on the ASC task.

\begin{table}
	\begin{minipage}[t]{0.5\linewidth}
		\centering
		\fontsize{10}{13}\selectfont
		\begin{tabular}{l|cccc}
			\hline
			{Dataset}&
			{\#Pos}&{\#Neg}&{\#Neu} & {\#Total}  \cr\hline
			Restaurant-Train&2164&807&637 &3608 \cr
			Restaurant-Test&728&196&196 &1120\cr
			\hline
			Laptop-Train&994&870&464 &2328\cr
			Laptop-Test&341&128&169 &638\cr\hline
		\end{tabular}
		\caption{Statistics of the ASC datasets.}
		\label{table:table1}
	\end{minipage}%
	\begin{minipage}[t]{0.5\linewidth}
		\centering
		\fontsize{10}{12}\selectfont
		\begin{tabular}{c|cccc}
			\hline
			{Datasets} & {\#Pos} & {\#Neg} & {\#Total} \\
			\hline
			Yelp Review & 266k & 177k & 443k \\
			Amazon Review & 277k & 277k & 554k \\
			\hline
		\end{tabular}
		\caption{Statistics of the DSC datasets.}
		\label{table:table2}
	\end{minipage}
\end{table}



\subsection{Experimental Settings}
In our experiments, word embeddings are initialized by 300-dimension GloVe~\cite{Pennington2014GloveGV}. After initialization, the word vectors are fixed and not fine-tuned during the training stage. All the weight matrices and biases are given the initial value by sampling from the uniform distribution $U(-0.1, 0.1)$. The dimension of LSTM cell hidden states is set to 300. We employ stochastic gradient descent (SGD) with momentum~\cite{DBLP:journals/nn/Qian99} to train models. The initial learning rate and momentum parameter are respectively set to 0.1 and 0.9. In addition, we apply dropout~\cite{DBLP:journals/corr/abs-1207-0580} with probability 0.5 on embedding layer as a regularizer. The parameter $\lambda$ in \emph{Attention Guidance} approach is set to 0.4. All hyper-parameters were tuned on 20\% randomly held-out training data. Finally, we run each model five times and report the average result of them.

\subsection{Compared Methods}
We divide compared methods into two groups according to whether using transferred knowledge. 

\noindent (I). The first group contains some classic methods for the ASC task:

\textbf{Majority} assigns each instance in the test set with the most frequent sentiment label in the training set. 

\textbf{Feature-based SVM}~\cite{kiritchenko2014nrc} is the top system of SemEval 2014 Task 4. It uses n-gram features, parse features and lexicon features to train an SVM classifier. 

\textbf{TD-LSTM}~\cite{Tang2016EffectiveLF} applies two LSTM networks to model the left context and right context of opinion target respectively, then concatenates their last hidden states for sentiment prediction. 

\textbf{ATAE-LSTM}~\cite{wang2016attention} concatenates the word embedding and target embedding as the input of LSTM, then employs the attention mechanism to capture target-dependent sentiment information. 

\textbf{IAN}~\cite{ma2017interactive} proposes the interactive attention to interactively learn representations of the context and target. The two representations are then concatenated for prediction. 

\textbf{MemNet}~\cite{Tang2016AspectLS} uses multi-hops attention on the word embeddings to generate the target-dependent sentence representation. 

\textbf{RAM}~\cite{chen2017recurrent} works similar to the method MemNet. It employs BiLSTM to build memory and applies GRU-based multi-hops attention. 

\textbf{IARM}~\cite{majumder2018iarm} incoporates the neighboring targets-related information for ASC by using memory networks.

\textbf{MGAN}~\cite{fan2018multi} proposes a fine-grained attention mechanism to capture the word-level interaction between target and context, then combines it with coarse-grained attention for ASC. 

\textbf{GCAE}~\cite{xue2018aspect} uses a convolutional neural network (CNN) with gating mechanisms to perform the ASC task. 

\textbf{TNet}~\cite{li2018transformation} proposes target specific transformation component to integrate target information into the word representation.

\noindent (II). Besides, we also compare two existing methods using transferred knowledge from large-scale DSC data to facilitate the ASC task:

\textbf{PRET+MULT}~\cite{he2018exploiting} shares shadow embedding and LSTM layers between the ASC model and the DSC model through multi-task learning. 

\textbf{TransCap}~\cite{chen2019transfer} employs capsule network to share the bottom features between the ASC task and the DSC task.




\begin{table}[t]
	\centering
	\fontsize{10}{12}\selectfont
	\setlength{\tabcolsep}{3.0mm}{
		\begin{tabular}{lcccc}
			\hline
			\multirow{2}*{Method}&\multicolumn{2}{c}{Restaurant}&\multicolumn{2}{c}{Laptop}\cr\cline{2-5}&Acc.&Macro-F1&Acc.&Macro-F1\cr\hline
			Majority&65.00&33.33&53.50&33.33\cr
			Feature-SVM~\cite{kiritchenko2014nrc}&80.16&N/A&70.49&N/A\cr
			ATAE-LSTM~\cite{wang2016attention}&77.20&N/A&68.70&N/A\cr
			TD-LSTM~\cite{Tang2016EffectiveLF}&78.00&66.73&71.83&68.43\cr
			IAN~\cite{ma2017interactive}&78.60&N/A&72.10&N/A\cr
			MemNet~\cite{Tang2016AspectLS}&80.32&N/A&72.37&N/A\cr
			RAM~\cite{chen2017recurrent}&80.23&70.80&74.49&71.35\cr
			IARM~\cite{majumder2018iarm}&80.00&N/A&73.80&N/A\cr
			MGAN~\cite{fan2018multi}&81.25&71.94&75.39&72.47\cr
			GCAE~\cite{xue2018aspect}&77.43&66.24&71.03&64.43\cr
			TNet~\cite{li2018transformation}&80.79&70.84&76.01&71.47\cr
			\hline
			PRET+MULT~\cite{he2018exploiting}&79.98&69.39&74.14&69.14\cr
			TransCap~\cite{chen2019transfer}&80.72&71.98&74.92&70.21\cr
			\hline
			Base ASC model&80.38&70.69&73.52&70.78\cr
			\textbf{ATN-AG}&$\text{81.39}^{\dagger}$&$\text{72.44}^{\dagger}$&$\text{76.41}^{\dagger}$&$\text{72.59}^{\dagger}$\cr
			\textbf{ATN-AF}&$\textbf{82.36}^{\dagger}$&$\textbf{74.00}^{\dagger}$&$\textbf{76.48}^{\dagger}$&$\textbf{72.60}^{\dagger}$
			\cr\hline
	\end{tabular}}
	\caption{Main experiment results (\%). The base ASC model is attention-based BiLSTM enhanced with position embedding. AT-AG and ATN-AF respectively refer to ATN model using \emph{Attention Guidance} and \emph{Attention Fusion}. The best performances are marked in bold. The marker ${\dagger}$ represents that ATN-AG and ATN-AF outperform the compared methods significantly (p $<$ 0.05).}
	\label{table:table3}
\end{table}

\subsection{Main Results and Analysis}
The main results are shown in Table~\ref{table:table3}. We classify the results into three groups: the first lists the classic methods for the ASC task, the second presents two existing transfer-based methods, and the last is our base ASC model and enhanced versions with transferring attention knowledge. We use ATN-AG and ATN-AF respectively to represent ATN using \emph{Attention Guidance} and \emph{Attention Fusion}.

The method Feature-SVM obtains competitive results on the restaurant dataset but performs poorly on the laptop dataset. This may be attributed to that the performance of simple feature-based methods heavily relies on the quality of hand-crafted features. IAN achieves better performance than TD-LSTM and ATAE-LSTM by using the interactive attention mechanism to learn the representations of context and opinion target. With combining of fine-grained and coarse-grained attention mechanisms, MGAN achieves the best performance among all pure attention-based models. Among the memory-based methods, it can be observed that RAM outperforms MemNet and IARM on the laptop dataset, which validates the effectiveness of multi-hops attention based on recurrent network. GCAE performs poorly compared with other neural methods, as CNN is not good at capturing the long-term dependencies between context words. TNet achieves state-of-the-art performance by designing target-specific transformation mechanism between LSTM and CNN.

PRET+MULT and Transcap transfer knowledge implicitly from large-scale DSC data to the ASC task through sharing parameters and multi-task learning. They show superiority compared to some methods without transferring knowledge. For example, the base model of PRET+MULT is an attention-based LSTM similar to ATAE-LSTM. We can observe that PRET+MULT outperforms ATAE-LSTM significantly, and achieves 2.78\% and 5.44\% accuracy improvements respectively on the restaurant and laptop datasets. Transcap obtains better results compared to PRET+MULT, which verifies the effectiveness of capsule network for capturing shared features.

Our base ASC model attention-based BiLSTM enhanced with position embedding performs better than some attention-based models, such as ATAE-LSTM and IAN. This result indicates that position embedding is beneficial for modeling target information in the ASC task. On this basis, our attention transfer models ATN-AG and ATN-AF respectively achieve about 1\% and 2\% improvements in accuracy on the restaurant dataset, and over 2.8\% improvements on the laptop dataset. In addition, they surpass two existing methods that use transferred knowledge obviously, i.e., PRET+MULT and Transcap. These comparisons demonstrate the effectiveness of our proposal of explicitly transferring attention knowledge from resource-rich DSC data to the ASC task. Compared with ATN-AG, ATN-AF achieves better performance on the restaurant dataset. It is reasonable because ATN-AG cannot leverage the attention weights of the DSC module during the testing stage. Nevertheless, ATN-AG still obtains comparable results on the laptop dataset and has a faster inference speed than ATN-AF.

\subsection{Effect of DSC Data Size}
To investigate the effect of DSC data size on our approaches, we vary the percentage of DSC data from 0\% to 100\% to report the results of ATN-AG and ATN-AF. The critical values 0\% and 100\% respectively mean no DSC data and using the complete DSC dataset. The results are shown in Figure~\ref{datasizefig}.

We can observe that our approaches ATN-AG and ATN-AF both achieve very stable improvements on the two datasets with the increase of DSC data size. This indicates that the ASC task indeed benefits from the transferred attention knowledge from the pre-trained DSC module. Consistent and stable improvements show the robustness of our approaches.

\subsection{Effect of Hyper-parameter $\lambda$}
\begin{figure}
	\begin{minipage}[t]{0.48\linewidth}
		\centering
		\label{fig:c}     
		\includegraphics[width=0.99\columnwidth]{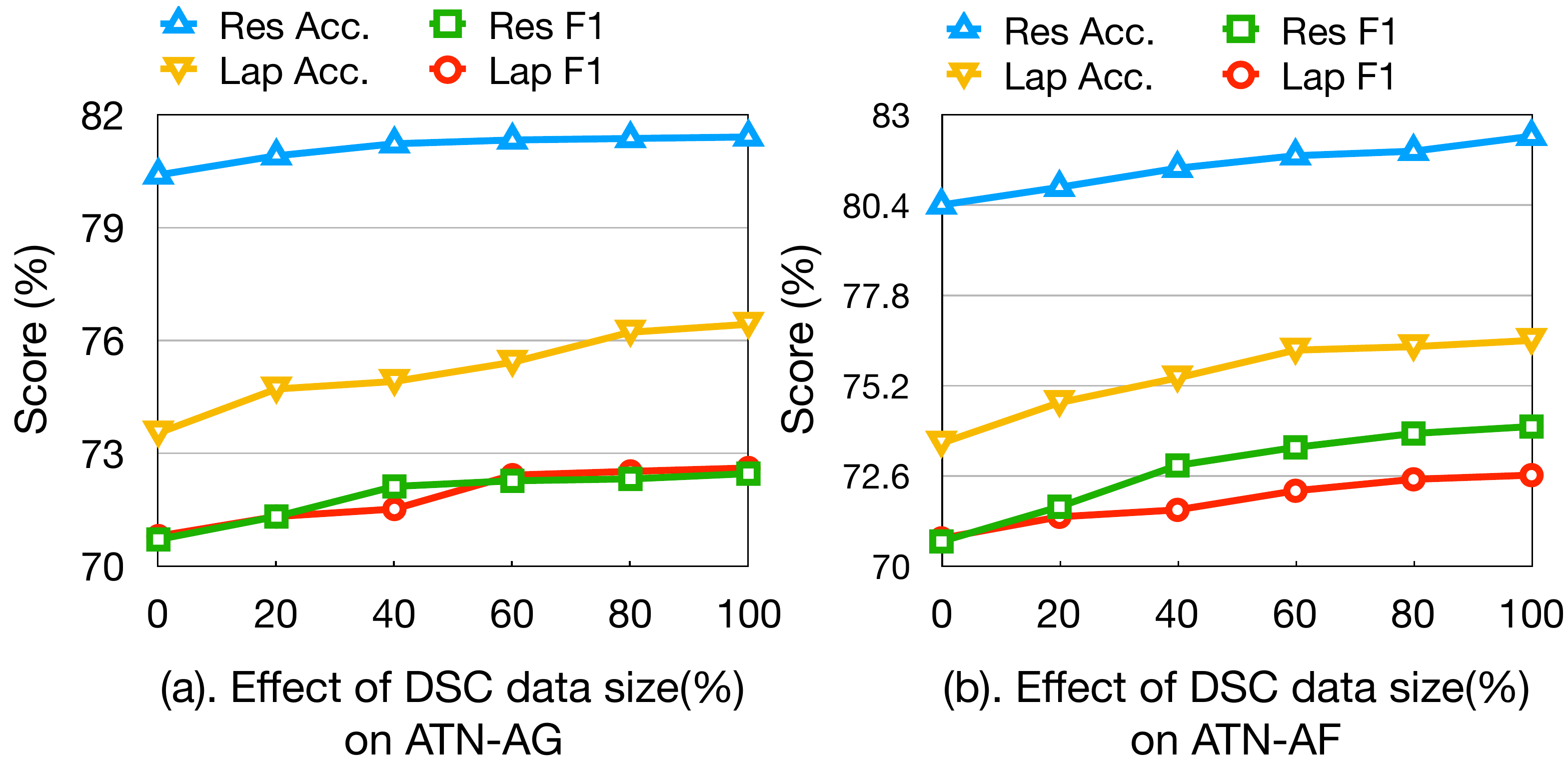}
		\caption{\centering Performance of ATN-AG and ATN-AF with different percentages of DSC data.}
		\label{datasizefig}
	\end{minipage}%
	\begin{minipage}[t]{0.52\linewidth}
		\centering
		\label{fig:b}     
		\includegraphics[width=0.99\columnwidth]{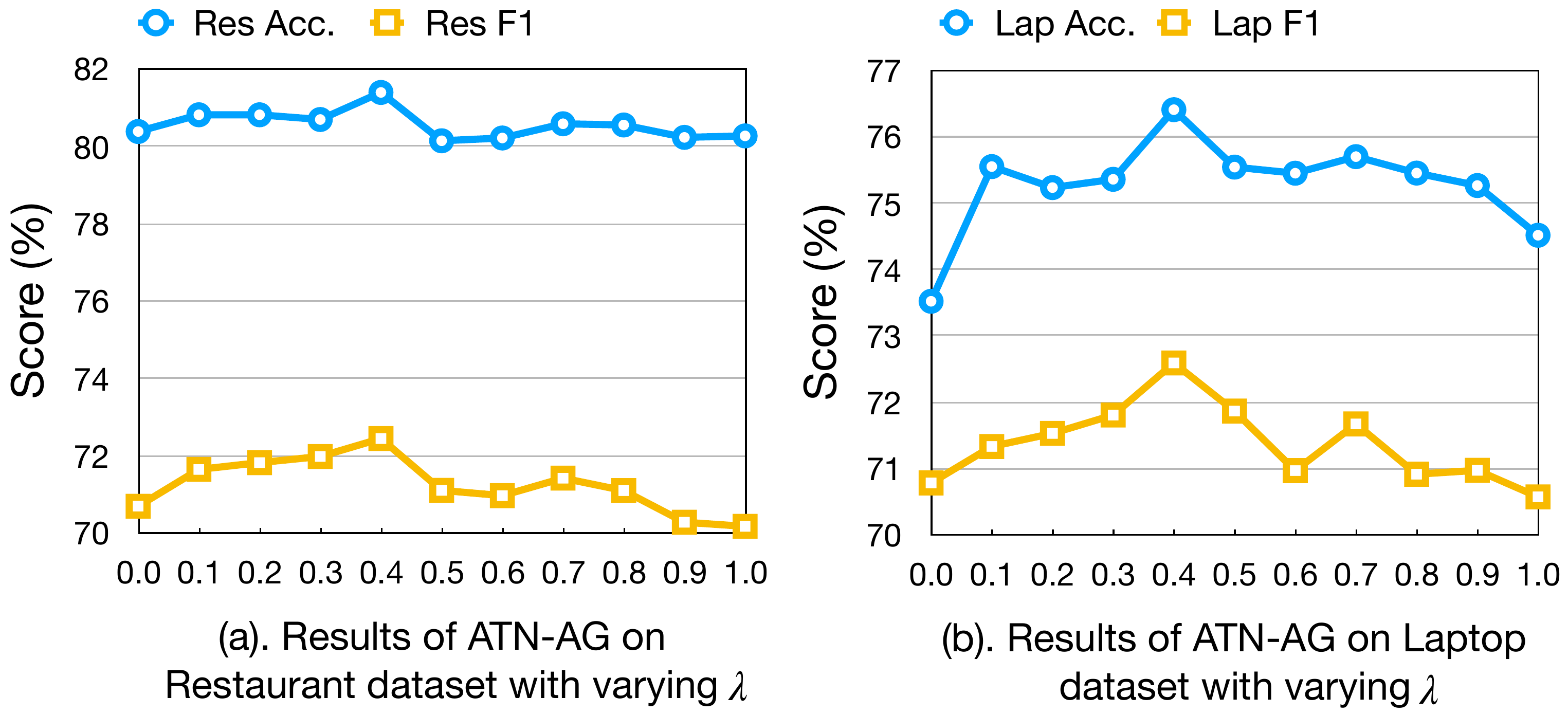} 
		\caption{ \centering Effect of hyper-parameter $\lambda$ on ATN-AG.}
		\label{lambdafig}
	\end{minipage}
\end{figure}

To analyze the effect of hyper-parameter $\lambda$ in Equation~\ref{lambda} on ATN-AG, we adjust it in [0, 1] to conduct experiments and the step is 0.1. Figure~\ref{lambdafig} shows the performance of ATN-AG with different $\lambda$ on the restaurant and laptop datasets.

We can see that the curves on two datasets have an overall upward trend when $\lambda$ $<$ 0.4, but become flat or downward once $\lambda$ $>$ 0.4. In the upward part, the attention knowledge from the DSC module is a useful guidance signal to help the ASC module to focus on sentiment words more accurately, thus improve the performance of ASC. Once the weight $\lambda$ exceeds 0.4, the transferred attention knowledge begins to dominate the attention process while the ASC module loses the mastership and perform worse. Therefore, we finally set $\lambda$ to be 0.4 on two datasets.

\begin{table}[t]
	\centering
	\fontsize{8.4}{15}\selectfont
	\begin{tabular}{p{1.5cm}|p{11cm}l}
		\hline
		Base model&{ I use \colorbox[RGB]{255,236,234}{it} \colorbox[RGB]{255,81,58}{mostly} for \textbf{[content creation]} ( Audio , video , photo editing ) and its reliable .}&Neagtive\xmark \cr	   
		ATN-AG&{I use it \colorbox[RGB]{255,241,239}{mostly} for \textbf{[content creation]} ( Audio , video , photo editing ) and its \colorbox{red}{reliable} .}&Positive\cmark \cr	
		ATN-AF&{I use \colorbox[RGB]{255,239,237}{it} mostly for \textbf{[content creation]} ( Audio , video , photo editing ) and \colorbox[RGB]{255,232,229}{its} \colorbox[RGB]{255,117,99}{reliable} .}&Positive\cmark \cr
		\hline

		Base model & \colorbox[RGB]{255,248,247}{Did} \colorbox[RGB]{255,236,234}{not} \colorbox[RGB]{255,74,50}{enjoy} the new Windows 8 and \textbf{[touchscreen functions]}&Positive\xmark \cr	   
		
		ATN-AG& \colorbox[RGB]{255,243,242}{Did} \colorbox[RGB]{255,106,86}{not} \colorbox[RGB]{255,209,204}{enjoy} the new Windows 8 and \textbf{[touchscreen functions]}&Negative\cmark \cr
		
		ATN-AF& \colorbox[RGB]{255,255,255}{Did} \colorbox[RGB]{255,122,104}{not} \colorbox[RGB]{255,198,191}{enjoy} the new Windows 8 and \textbf{[touchscreen functions]}&Negative\cmark \cr
		\hline
	\end{tabular}
	\caption{Attention visualization of ATN-AG and ATN-AF. The spans in bold are opinion targets. A darker color indicates a higher attention weight.}
	\label{attentionfig}
\end{table}

%

\subsection{Case Study}
In the ATN model, we propose the approaches \emph{Attention Guidance} and  \emph{Attention Fusion} to help the ASC module to capture sentiment clues more accurately. To verify this, we analyze some dozens of instances from the test set. Compared with the base ASC model, we find that our attention transfer methods can deal with low-frequency sentiment words and complex sentiment patterns such as negation. Table~\ref{attentionfig} shows the attention visualizations of two examples and the corresponding sentiment predictions under the base model, ATN-AG and ATN-AF. Note that the darker color means higher attention weight. 

In the first example, the base ASC model mainly focuses on the adverb ``\emph{mostly}'', while fails to capture the critical sentiment clue ``\emph{reliable}''. According to the statistics, the word ``\emph{reliable}'' only appears five times in the training set. This indicates that the base model is not good at catching low-frequency sentiment words, thus makes wrong sentiment predictions. In contrast, the enhanced models ATN-AG and ATN-AF with transferred attention knowledge both successfully capture the informative word ``\emph{reliable}'', and give the right predictions.

From the second example, we can see that the base ASC model mainly focuses on the word ``\emph{enjoy}'' rather than the sentiment negator ``\emph{not}''. It is hard for the base model to learn the negation with the insufficient labeled dataset. With the help of the external attention knowledge, our approaches ATN-AG and ATN-AF pay more attention to the negator ``\emph{not}'', and make correct sentiment predictions.

The above observations show that our approaches indeed improve the low-resource task ASC with the transferred attention knowledge and retain good interpretability.

\section{Related Work}
\subsection{Aspect-level Sentiment Classification}
Early works adopt supervised learning and devote to designing effective features for the ASC task, such as n-gram features~\cite{kiritchenko2014nrc} and sentiment lexicons~\cite{vo2015target}. The performance of these methods heavily depends on labor-intensive feature engineering. With the development of deep learning,~\newcite{Tang2016EffectiveLF} use two Long Short-Term Memory (LSTM)~\cite{hochreiter1997long} networks to respectively model the left context and right context of the given opinion target. However, it cannot capture the association between the context and opinion target. To address the issue, recent works employ the attention mechanism to catch target-dependent sentiment context and achieve very promising resutls~\cite{wang2016attention,ma2017interactive,fan2018multi}. Instead of single attention, some works propose multi-hops attention based on memory networks~\cite{sukhbaatar2015end} to detect more powerful sentiment clues~\cite{Tang2016AspectLS,chen2017recurrent,majumder2018iarm}.

Despite attention-based models showing the potential for ASC, they highly rely on data-driven attention mechanism. Unfortunately, public ASC datasets are all small-scale because of the complexity of annotation. Insufficient labeled data finally limits the effectiveness of attention mechanism for the ASC task. Different from the above methods, we improve the attention capacity of the ASC model in this work, by transferring substantial attention knowledge from the DSC model pre-trained with resource-rich document-level sentiment classification data.

\subsection{Transfer Learning}
Transfer learning aims to extract knowledge from one or more source tasks and then apply them to a target task. Neural transfer learning has proven effective for image recognition~\cite{donahue2014decaf} and natural language processing tasks~\cite{mou2016transferable,dong2018helping,wu2020latent}.~\newcite{he2018exploiting} are the first to transfer knowledge from document-level review data to improve the ASC task through sharing embedding and LSTM layers.~\newcite{chen2019transfer} employ capsule network to share bottom features between the ASC task and DSC task. In this work, we aim to transfer attention knowledge from the DSC model explicitly to improve the effectiveness of attention mechanism for the ASC task. In contrast to the two existing works, our proposed approaches show better performance and good interpretability.

\section{Conclusion}
Insufficient labeled data limits the effectiveness of attention-based models for the ASC task. In this paper, we propose a novel attention transfer framework, in which two different attention transfer methods are designed to exploit attention knowledge from resource-rich document-level sentiment classification corpus to enhance the attention process of resource-poor aspect-level sentiment classification, finally achieving the goal of improving the performance of ASC. Experimental results indicate that our approaches outperform the state-of-the-art works. Further analysis validates the effectiveness and benefits of transferring the attention knowledge from DSC data for the ASC task.

\section*{Acknowledgements}
We would like to thank the anonymous reviewers for their insightful comments. This work was supported by the NSFC (No. 61976114, 61936012) and National Key R\&D Program of China (No. 2018YFB1005102).

\bibliographystyle{coling}
\bibliography{coling2020}

\begin{thebibliography}{}

\bibitem[\protect\citename{Bahdanau \bgroup et al.\egroup
  }2014]{bahdanau2014neural}
Dzmitry Bahdanau, Kyunghyun Cho, and Yoshua Bengio.
\newblock 2014.
\newblock Neural machine translation by jointly learning to align and
  translate.
\newblock {\em arXiv preprint arXiv:1409.0473}.

\bibitem[\protect\citename{Bengio \bgroup et al.\egroup
  }2003]{bengio2003neural}
Yoshua Bengio, R{\'e}jean Ducharme, Pascal Vincent, and Christian Jauvin.
\newblock 2003.
\newblock A neural probabilistic language model.
\newblock {\em Journal of machine learning research}, 3(Feb):1137--1155.

\bibitem[\protect\citename{Chen and Qian}2019]{chen2019transfer}
Zhuang Chen and Tieyun Qian.
\newblock 2019.
\newblock Transfer capsule network for aspect level sentiment classification.
\newblock In {\em Proceedings of the 57th Annual Meeting of the Association for
  Computational Linguistics}, pages 547--556.

\bibitem[\protect\citename{Chen \bgroup et al.\egroup }2017]{chen2017recurrent}
Peng Chen, Zhongqian Sun, Lidong Bing, and Wei Yang.
\newblock 2017.
\newblock Recurrent attention network on memory for aspect sentiment analysis.
\newblock In {\em Proceedings of the 2017 conference on empirical methods in
  natural language processing}, pages 452--461.

\bibitem[\protect\citename{Dahl \bgroup et al.\egroup
  }2012]{DBLP:journals/taslp/DahlYDA12}
George~E. Dahl, Dong Yu, Li~Deng, and Alex Acero.
\newblock 2012.
\newblock Context-dependent pre-trained deep neural networks for
  large-vocabulary speech recognition.
\newblock {\em {IEEE} Trans. Audio, Speech {\&} Language Processing},
  20(1):30--42.

\bibitem[\protect\citename{Donahue \bgroup et al.\egroup
  }2014]{donahue2014decaf}
Jeff Donahue, Yangqing Jia, Oriol Vinyals, Judy Hoffman, Ning Zhang, Eric
  Tzeng, and Trevor Darrell.
\newblock 2014.
\newblock Decaf: A deep convolutional activation feature for generic visual
  recognition.
\newblock In {\em International conference on machine learning}, pages
  647--655.

\bibitem[\protect\citename{Dong and De~Melo}2018]{dong2018helping}
Xin Dong and Gerard De~Melo.
\newblock 2018.
\newblock A helping hand: Transfer learning for deep sentiment analysis.
\newblock In {\em Proceedings of the 56th Annual Meeting of the Association for
  Computational Linguistics (Volume 1: Long Papers)}, pages 2524--2534.

\bibitem[\protect\citename{Fan \bgroup et al.\egroup }2018]{fan2018multi}
Feifan Fan, Yansong Feng, and Dongyan Zhao.
\newblock 2018.
\newblock Multi-grained attention network for aspect-level sentiment
  classification.
\newblock In {\em Proceedings of the 2018 Conference on Empirical Methods in
  Natural Language Processing}, pages 3433--3442.

\bibitem[\protect\citename{Gehring \bgroup et al.\egroup
  }2016]{gehring2016convolutional}
Jonas Gehring, Michael Auli, David Grangier, and Yann~N Dauphin.
\newblock 2016.
\newblock A convolutional encoder model for neural machine translation.
\newblock {\em arXiv preprint arXiv:1611.02344}.

\bibitem[\protect\citename{He \bgroup et al.\egroup }2018]{he2018exploiting}
Ruidan He, Wee~Sun Lee, Hwee~Tou Ng, and Daniel Dahlmeier.
\newblock 2018.
\newblock Exploiting document knowledge for aspect-level sentiment
  classification.
\newblock {\em arXiv preprint arXiv:1806.04346}.

\bibitem[\protect\citename{Hinton \bgroup et al.\egroup
  }2012]{DBLP:journals/corr/abs-1207-0580}
Geoffrey~E. Hinton, Nitish Srivastava, Alex Krizhevsky, Ilya Sutskever, and
  Ruslan Salakhutdinov.
\newblock 2012.
\newblock Improving neural networks by preventing co-adaptation of feature
  detectors.
\newblock {\em CoRR}, abs/1207.0580.

\bibitem[\protect\citename{Hochreiter and Schmidhuber}1997]{hochreiter1997long}
Sepp Hochreiter and J{\"u}rgen Schmidhuber.
\newblock 1997.
\newblock Long short-term memory.
\newblock {\em Neural computation}, 9(8):1735--1780.

\bibitem[\protect\citename{Jiang \bgroup et al.\egroup }2011]{jiang2011target}
Long Jiang, Mo~Yu, Ming Zhou, Xiaohua Liu, and Tiejun Zhao.
\newblock 2011.
\newblock Target-dependent twitter sentiment classification.
\newblock In {\em Proceedings of the 49th Annual Meeting of the Association for
  Computational Linguistics: Human Language Technologies-Volume 1}, pages
  151--160. Association for Computational Linguistics.

\bibitem[\protect\citename{Kiritchenko \bgroup et al.\egroup
  }2014]{kiritchenko2014nrc}
Svetlana Kiritchenko, Xiaodan Zhu, Colin Cherry, and Saif Mohammad.
\newblock 2014.
\newblock Nrc-canada-2014: Detecting aspects and sentiment in customer reviews.
\newblock In {\em Proceedings of the 8th International Workshop on Semantic
  Evaluation (SemEval 2014)}, pages 437--442.

\bibitem[\protect\citename{Krizhevsky \bgroup et al.\egroup
  }2012]{krizhevsky2012imagenet}
Alex Krizhevsky, Ilya Sutskever, and Geoffrey~E Hinton.
\newblock 2012.
\newblock Imagenet classification with deep convolutional neural networks.
\newblock In {\em Advances in neural information processing systems}, pages
  1097--1105.

\bibitem[\protect\citename{Li \bgroup et al.\egroup }2018a]{li2018delete}
Juncen Li, Robin Jia, He~He, and Percy Liang.
\newblock 2018a.
\newblock Delete, retrieve, generate: A simple approach to sentiment and style
  transfer.
\newblock {\em arXiv preprint arXiv:1804.06437}.

\bibitem[\protect\citename{Li \bgroup et al.\egroup
  }2018b]{li2018transformation}
Xin Li, Lidong Bing, Wai Lam, and Bei Shi.
\newblock 2018b.
\newblock Transformation networks for target-oriented sentiment classification.
\newblock {\em arXiv preprint arXiv:1805.01086}.

\bibitem[\protect\citename{Lin \bgroup et al.\egroup }2016]{lin2016neural}
Yankai Lin, Shiqi Shen, Zhiyuan Liu, Huanbo Luan, and Maosong Sun.
\newblock 2016.
\newblock Neural relation extraction with selective attention over instances.
\newblock In {\em Proceedings of the 54th Annual Meeting of the Association for
  Computational Linguistics (Volume 1: Long Papers)}, pages 2124--2133.

\bibitem[\protect\citename{Liu}2012]{liu2012sentiment}
Bing Liu.
\newblock 2012.
\newblock Sentiment analysis and opinion mining.
\newblock {\em Synthesis lectures on human language technologies}, 5(1):1--167.

\bibitem[\protect\citename{Ma \bgroup et al.\egroup }2017]{ma2017interactive}
Dehong Ma, Sujian Li, Xiaodong Zhang, and Houfeng Wang.
\newblock 2017.
\newblock Interactive attention networks for aspect-level sentiment
  classification.
\newblock {\em arXiv preprint arXiv:1709.00893}.

\bibitem[\protect\citename{Majumder \bgroup et al.\egroup
  }2018]{majumder2018iarm}
Navonil Majumder, Soujanya Poria, Alexander Gelbukh, Md~Shad Akhtar, Erik
  Cambria, and Asif Ekbal.
\newblock 2018.
\newblock Iarm: Inter-aspect relation modeling with memory networks in
  aspect-based sentiment analysis.
\newblock In {\em Proceedings of the 2018 conference on empirical methods in
  natural language processing}, pages 3402--3411.

\bibitem[\protect\citename{Mou \bgroup et al.\egroup
  }2016]{mou2016transferable}
Lili Mou, Zhao Meng, Rui Yan, Ge~Li, Yan Xu, Lu~Zhang, and Zhi Jin.
\newblock 2016.
\newblock How transferable are neural networks in nlp applications?
\newblock {\em arXiv preprint arXiv:1603.06111}.

\bibitem[\protect\citename{Pang \bgroup et al.\egroup }2008]{pang2008opinion}
Bo~Pang, Lillian Lee, et~al.
\newblock 2008.
\newblock Opinion mining and sentiment analysis.
\newblock {\em Foundations and Trends{\textregistered} in Information
  Retrieval}, 2(1--2):1--135.

\bibitem[\protect\citename{Pennington \bgroup et al.\egroup
  }2014]{Pennington2014GloveGV}
Jeffrey Pennington, Richard Socher, and Christopher~D. Manning.
\newblock 2014.
\newblock Glove: Global vectors for word representation.
\newblock In {\em EMNLP}.

\bibitem[\protect\citename{Pontiki \bgroup et al.\egroup
  }2014]{Pontiki2014SemEval2014T4}
Maria Pontiki, Dimitris Galanis, John Pavlopoulos, Harris Papageorgiou, Ion
  Androutsopoulos, and Suresh Manandhar.
\newblock 2014.
\newblock Semeval-2014 task 4: Aspect based sentiment analysis.
\newblock In {\em COLING 2014}.

\bibitem[\protect\citename{Qian}1999]{DBLP:journals/nn/Qian99}
Ning Qian.
\newblock 1999.
\newblock On the momentum term in gradient descent learning algorithms.
\newblock {\em Neural Networks}, 12(1):145--151.

\bibitem[\protect\citename{Sabour \bgroup et al.\egroup
  }2017]{sabour2017dynamic}
Sara Sabour, Nicholas Frosst, and Geoffrey~E Hinton.
\newblock 2017.
\newblock Dynamic routing between capsules.
\newblock In {\em Advances in neural information processing systems}, pages
  3856--3866.

\bibitem[\protect\citename{Sukhbaatar \bgroup et al.\egroup
  }2015]{sukhbaatar2015end}
Sainbayar Sukhbaatar, Jason Weston, Rob Fergus, et~al.
\newblock 2015.
\newblock End-to-end memory networks.
\newblock In {\em Advances in neural information processing systems}, pages
  2440--2448.

\bibitem[\protect\citename{Tang \bgroup et al.\egroup
  }2016a]{Tang2016EffectiveLF}
Duyu Tang, Bing Qin, Xiaocheng Feng, and Ting Liu.
\newblock 2016a.
\newblock Effective lstms for target-dependent sentiment classification.
\newblock In {\em COLING}.

\bibitem[\protect\citename{Tang \bgroup et al.\egroup }2016b]{Tang2016AspectLS}
Duyu Tang, Bing Qin, and Ting Liu.
\newblock 2016b.
\newblock Aspect level sentiment classification with deep memory network.
\newblock In {\em EMNLP}.

\bibitem[\protect\citename{Vo and Zhang}2015]{vo2015target}
Duy-Tin Vo and Yue Zhang.
\newblock 2015.
\newblock Target-dependent twitter sentiment classification with rich automatic
  features.
\newblock In {\em Twenty-Fourth International Joint Conference on Artificial
  Intelligence}.

\bibitem[\protect\citename{Wang \bgroup et al.\egroup }2016]{wang2016attention}
Yequan Wang, Minlie Huang, Li~Zhao, et~al.
\newblock 2016.
\newblock Attention-based lstm for aspect-level sentiment classification.
\newblock In {\em Proceedings of the 2016 conference on empirical methods in
  natural language processing}, pages 606--615.

\bibitem[\protect\citename{Wu \bgroup et al.\egroup }2020]{wu2020latent}
Zhen Wu, Fei Zhao, Xin-Yu Dai, Shujian Huang, and Jiajun Chen.
\newblock 2020.
\newblock Latent opinions transfer network for target-oriented opinion words
  extraction.
\newblock {\em arXiv preprint arXiv:2001.01989}.

\bibitem[\protect\citename{Xue and Li}2018]{xue2018aspect}
Wei Xue and Tao Li.
\newblock 2018.
\newblock Aspect based sentiment analysis with gated convolutional networks.
\newblock {\em arXiv preprint arXiv:1805.07043}.

\end{thebibliography}

\end{document}